\documentclass[]{article}
\usepackage{proceed2e}

\usepackage{amsmath}
\usepackage{amssymb}
\usepackage{graphicx}
\usepackage{enumitem}
\usepackage{subfigure}
\usepackage{rpa-misc}
\usepackage{rpa-algorithm}

\title{Bayesian Online Changepoint Detection}
\author{
  \bf{Ryan Prescott Adams}\\
  Cavendish Laboratory\\
  Cambridge CB3 0HE\\
  United Kingdom
  \And
  \bf{David J.C. MacKay}\\
  Cavendish Laboratory\\
  Cambridge CB3 0HE\\
  United Kingdom
}

\newcommand{\boldeta}{\ensuremath{\boldsymbol{\eta}}}
\newcommand{\bchi}{\ensuremath{\boldsymbol{\chi}}}

\begin{document}
  \maketitle

  \begin{abstract}
    Changepoints are abrupt variations in the generative parameters of a
    data sequence.  Online detection of changepoints is useful in modelling
    and prediction of time series in application areas such as finance,
    biometrics, and robotics.  While frequentist methods have yielded
    online filtering and prediction techniques, most Bayesian papers have
    focused on the retrospective segmentation problem.  Here we examine the
    case where the model parameters before and after the changepoint are
    independent and we derive an online algorithm for exact inference of
    the most recent changepoint.  We compute the probability distribution
    of the length of the current ``run,'' or time since the last
    changepoint, using a simple message-passing algorithm.  Our
    implementation is highly modular so that the algorithm may be applied
    to a variety of types of data.  We illustrate this modularity by
    demonstrating the algorithm on three different real-world data sets.
  \end{abstract}

  \section{INTRODUCTION}
    Changepoint detection is the identification of abrupt changes in the
    generative parameters of sequential data.  As an online and offline
    signal processing tool, it has proven to be useful in applications such
    as process control \cite{aroian-levene-1950a}, EEG analysis
    \cite{bodenstein-praetorius-1977a, barlow-etal-1981a,
    kaplan-shishkin-2000a}, DNA segmentation \cite{braun-etal-2000a},
    econometrics \cite{chen-gupta-1997a, koop-potter-2004a}, and disease
    demographics \cite{denison-holmes-1999a}.

    Frequentist approaches to changepoint detection, from the pioneering
    work of Page \cite{page-1954a, page-1955a} and Lorden
    \cite{lorden-1971a} to recent work using support vector machines
    \cite{desobry-etal-2005a}, offer online changepoint detectors.  Most
    Bayesian approaches to changepoint detection, in contrast, have been
    offline and retrospective \cite{smith-1975a, barry-hartigan-1993a,
    stephens-1994a, green-1995a, chib-1998a}.  With a few exceptions
    \cite{jervis-jardine-1997a, oruanaidh-etal-1994a}, the Bayesian papers
    on changepoint detection focus on segmentation and techniques to
    generate samples from the posterior distribution over changepoint
    locations.

    In this paper, we present a Bayesian changepoint detection algorithm
    for online inference.  Rather than retrospective segmentation, we focus
    on causal predictive filtering; generating an accurate distribution of
    the next unseen datum in the sequence, given only data already
    observed.  For many applications in machine intelligence, this is a
    natural requirement.  Robots must navigate based on past sensor data
    from an environment that may have abruptly changed: a door may be
    closed now, for example, or the furniture may have been moved.  In
    vision systems, the brightness change when a light switch is flipped or
    when the sun comes out.

    We assume that a sequence of observations $x_{1},x_{2},\ldots,x_{T}$
    may be divided into non-overlapping \textit{product partitions}
    \cite{barry-hartigan-1992a}.  The delineations between partitions are
    called the changepoints.  We further assume that for each partition
    $\rho$, the data within it are i.i.d.\ from some probability
    distribution $P(x_{t} \given \boldeta_{\rho})$.  The parameters
    $\boldeta_{\rho}$, $\rho = 1,2,\ldots$ are taken to be i.i.d.\ as well.
    We denote the contiguous set of observations between time $a$ and $b$
    inclusive as $\bx_{a:b}$.  The discrete \textit{a priori} probability
    distribution over the interval between changepoints is denoted as
    $P_{\sf{gap}}(g)$.

\begin{figure}[!t]
      \centering 
      \subfigure[]{\label{fig:example-model-a}
	\includegraphics{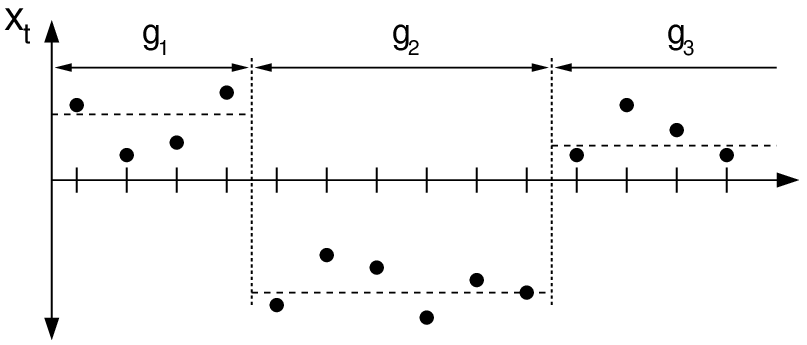}}\\
      \subfigure[]{\label{fig:example-model-b}
	\includegraphics{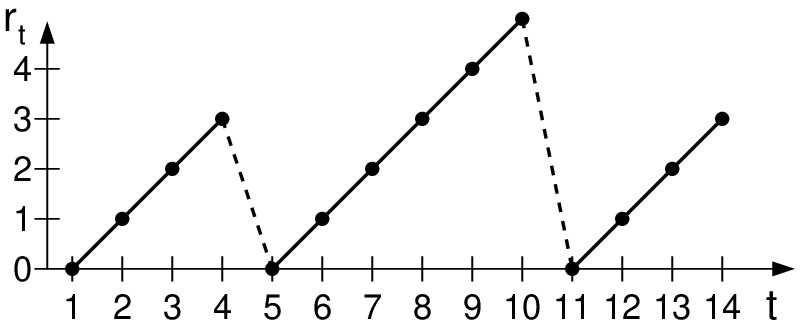}}\\
      \subfigure[]{\label{fig:example-model-c}
	\includegraphics{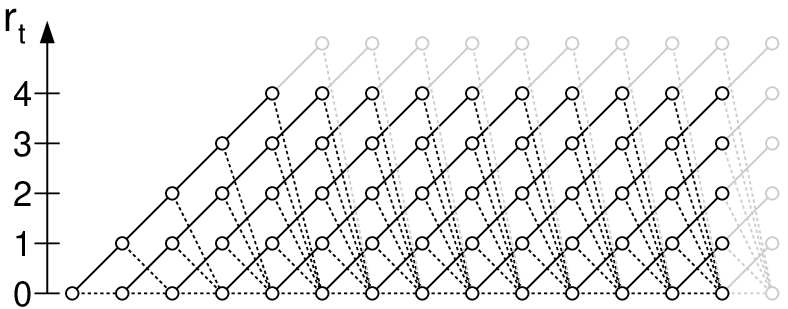}}\\
      \caption{This figure illustrates how we describe a changepoint model
      expressed in terms of run lengths.  Figure \ref{fig:example-model-a}
      shows hypothetical univariate data divided by changepoints on the
      mean into three segments of lengths $g_{1}=4$, $g_{2}=6$, and an
      undetermined length $g_{3}$.  Figure \ref{fig:example-model-b} shows
      the run length $r_{t}$ as a function of time. $r_{t}$ drops to zero
      when a changepoint occurs.  Figure \ref{fig:example-model-c} shows
      the trellis on which the message-passing algorithm lives.  Solid
      lines indicate that probability mass is being passed ``upwards,''
      causing the run length to grow at the next time step.  Dotted lines
      indicate the possibility that the current run is truncated and the
      run length drops to zero.}
      \label{fig:example-model}
    \end{figure}

    We are concerned with estimating the posterior distribution over the
    current ``run length,'' or time since the last changepoint, given the
    data so far observed.  We denote the length of the current run at time
    $t$ by $r_{t}$.  We also use the notation $\bx_{t}^{(r)}$ to indicate
    the set of observations associated with the run $r_{t}$.  As $r$ may be
    zero, the set $\bx^{(r)}$ may be empty.  We illustrate the relationship
    between the run length $r$ and some hypothetical univariate data in
    Figures \ref{fig:example-model-a} and \ref{fig:example-model-b}.

  \section{RECURSIVE RUN LENGTH ESTIMATION}
    We assume that we can compute the predictive distribution conditional
    on a given run length $r_{t}$.  We then integrate over the posterior
    distribution on the current run length to find the marginal predictive
    distribution:
    \begin{align}\label{eqn:predictive}
      P(x_{t+1} \given \bx_{1:t}) &=
      \sum_{r_{t}} P(x_{t+1} \given r_{t}, \bx_{t}^{(r)})
                   P(r_{t} \given \bx_{1:t})
    \end{align}

    To find the posterior distribution
    \begin{align}\label{eqn:run-length-posterior}
      P(r_{t} \given \bx_{1:t}) &=
      \frac{P(r_{t},\bx_{1:t})}{P(\bx_{1:t})},
    \end{align}
    we write the joint distribution over run length and observed data
    recursively.
    \begin{align}
      P(&r_{t},\bx_{1:t}) = \sum_{r_{t-1}} P(r_{t},r_{t-1},\bx_{1:t})\notag\\
      \label{eqn:recursive-joint-1}
      &= \sum_{r_{t-1}} P(r_{t},x_{t} \given r_{t-1}, \bx_{1:t-1})
                        P(r_{t-1}, \bx_{1:t-1})\\
      &= \sum_{r_{t-1}} P(r_{t} \given r_{t-1})
                        P(x_{t} \given r_{t-1}, \bx_{t}^{(r)})
			P(r_{t-1},\bx_{1:t-1})\notag
    \end{align}
    Note that the predictive distribution $P(x_{t}\given r_{t-1},
    \bx_{1:t})$ depends only on the recent data $\bx_{t}^{(r)}$.  We can
    thus generate a recursive message-passing algorithm for the joint
    distribution over the current run length and the data, based on two
    calculations: 1)~the prior over $r_{t}$ given $r_{t-1}$, and 2)~the
    predictive distribution over the newly-observed datum, given the data
    since the last changepoint.

    \subsection{THE CHANGEPOINT PRIOR}
      The conditional prior on the changepoint $P(r_{t} \given r_{t-1})$
      gives this algorithm its computational efficiency, as it has nonzero
      mass at only two outcomes: the run length either continues to grow
      and $r_{t} = r_{t-1}+1$ or a changepoint occurs and $r_{t}=0$.
      \begin{align}\label{eqn:cond-prior}
	P(r_{t}\given r_{t-1}) = \left\{\begin{array}{ll}
	  H(r_{t-1}\!+\!1) & \mbox{if $r_{t}=0$}\\
	  1 - H(r_{t-1}\!+\!1) & \mbox{if $r_{t}=r_{t-1}+1$}\\
	  0 & \mbox{otherwise}
	\end{array}\right.
      \end{align}
      The function $H(\tau)$ is the \textit{hazard function}.
      \cite{evans-etal-2000a}.
      \begin{align}\label{eqn:hazard-function}
	H(\tau) &= \frac{P_{\sf{gap}}(g\shorteq\tau)}
	                {\sum_{t=\tau}^{\infty}P_{\sf{gap}}(g\shorteq t)}
      \end{align}
      In the special case is where $P_{\sf{gap}}(g)$ is a discrete
      exponential (geometric) distribution with timescale $\lambda$, the
      process is memoryless and the hazard function is constant at
      $H(\tau)=1/\lambda$.

      Figure \ref{fig:example-model-c} illustrates the resulting
      message-passing algorithm.  In this diagram, the circles represent
      run-length hypotheses.  The lines between the circles show recursive
      transfer of mass between time steps.  Solid lines indicate that
      probability mass is being passed ``upwards,'' causing the run length
      to grow at the next time step.  Dotted lines indicate that the
      current run is truncated and the run length drops to zero.
      
    \subsection{BOUNDARY CONDITIONS}
      A recursive algorithm must not only define the recurrence relation,
      but also the initialization conditions.  We consider two cases: 1)~a
      changepoint occurred \textit{a priori} before the first datum, such
      as when observing a game.  In such cases we place all of the
      probability mass for the initial run length at zero, i.e.\
      $P(r_{0}\shorteq 0) = 1$.  2)~We observe some recent subset of the
      data, such as when modelling climate change.  In this case the prior
      over the initial run length is the normalized \textit{survival
      function} \cite{evans-etal-2000a}
      \begin{align}\label{eqn:survival-function}
	P(r_{0}\shorteq \tau) &= \frac{1}{Z}S(\tau),
      \end{align}
      where $Z$ is an appropriate normalizing constant, and
      \begin{align}
	S(\tau) &= \sum_{t=\tau+1}^{\infty}P_{\sf{gap}}(g\shorteq t).
      \end{align}
    
    \subsection{CONJUGATE-EXPONENTIAL MODELS}
      Conjugate-exponential models are particularly convenient for
      integrating with the changepoint detection scheme described here.
      Exponential family likelihoods allow inference with a finite number
      of sufficient statistics which can be calculated incrementally as
      data arrives.  Exponential family likelihoods have the form
      \begin{align}\label{eqn:exponential-likelihood}
        P(\bx \given \boldeta) = h(\bx)\exp\left(\boldeta^{\intercal}
                                 \boldsymbol{U}(\bx)
	                         - A(\boldeta)\right)
      \end{align}
      where
      \begin{align}\label{eqn:exponential-normalizer}
	A(\boldeta) &= \log\int d\boldeta\;
	h(\bx)\exp\left(\boldeta^{\intercal}\boldsymbol{U}(\bx)\right).
      \end{align}

          \begin{algorithm}[t]
      \fbox{\begin{minipage}[t]{3.2in}
      \begin{enumerate}[itemsep=0cm]
	\item \textbf{Initialize}
	  \vskip -0.8cm
	  \begin{equation*}
	    \begin{split}
	      P(r_{0}) = \tilde{S}(r) &\text{ or } P(r_{0}\shorteq 0)=1\\
	      \nu_{1}^{(0)} &= \nu_{\sf{prior}}\\
	      \bchi_{1}^{(0)} &= \bchi_{\sf{prior}}\\[-0.3cm]
	    \end{split}
	  \end{equation*}
	\item \textbf{Observe New Datum $x_{t}$}
	\item \textbf{Evaluate Predictive Probability}
	  \vskip -0.5cm
	  \begin{equation*}
	    \pi_{t}^{(r)} =
	    P(x_{t}\given \nu_{t}^{(r)},\bchi_{t}^{(r)})\\[-0.2cm]
	  \end{equation*}
	\item \textbf{Calculate Growth Probabilities}
	  \vskip -0.8cm
	  \begin{equation*}
	      P(r_{t}\shorteq r_{t\!-\!1}\!+\!1,\bx_{1:t}) =
	      P(r_{t\!-\!1},\!\bx_{1:t\!-\!1})\pi_{t}^{(r)}(1\!-\!H(r_{t\!-\!1}))
	      \\[-0.2cm]
	  \end{equation*}	      
	\item \textbf{Calculate Changepoint Probabilities}
	  \vskip -0.8cm
	  \begin{equation*}
	      P(r_{t}\shorteq 0,\bx_{1:t}) =
	      \sum_{r_{t\!-\!1}}P(r_{t\!-\!1},\bx_{1:t\!-\!1})
	      \pi_{t}^{(r)}H(r_{t\!-\!1})\\[-0.3cm]
	  \end{equation*}
	\item \textbf{Calculate Evidence}
	  \vskip -0.5cm
	  \begin{equation*}
	    P(\bx_{1:t}) = \sum_{r_{t}}P(r_{t},\bx_{1:t})\\[-0.3cm]
	  \end{equation*}
	\item \textbf{Determine Run Length Distribution}
	  \vskip -0.5cm
	  \begin{equation*}
	   P(r_{t}\given \bx_{1:t}) = P(r_{t},\bx_{1:t})/
	   P(\bx_{1:t})\\[-0.2cm]
	  \end{equation*}
	\item \textbf{Update Sufficient Statistics}
	  \vskip -0.8cm
	  \begin{equation*}
	    \begin{split}
	    \nu_{t\!+\!1}^{(0)} &= \nu_{\sf{prior}}\\
	    \bchi_{t\!+\!1}^{(0)} &= \bchi_{\sf{prior}}\\
	    \nu_{t\!+\!1}^{(r\!+\!1)} &= \nu_{t}^{(r)} + 1\\
	    \bchi_{t\!+\!1}^{(r\!+\!1)} &= \bchi_{t}^{(r)} +
	    \boldsymbol{u}(x_{t})\\[-0.3cm]
	    \end{split}
	  \end{equation*}
	\item \textbf{Perform Prediction}
	  \vskip -0.8cm
	  \begin{equation*}
	    P(x_{t\!+\!1}\given\bx_{1:t}) =
	    \sum_{r_{t}}P(\bx_{t\!+\!1}|
	                      \bx_{t}^{(r)}\!,r_{t})
		              P(r_{t}|\bx_{1:t})\\[-0.4cm]
	  \end{equation*}
	\item \textbf{Return to Step 2}
      \end{enumerate}
      \end{minipage}}
      \caption{The online changepoint algorithm with prediction.  An
      additional optimization not shown is to truncate the per-timestep
      vectors when the tail of $P(r_{t}|\bx_{1:t})$ has mass beneath a
      threshold.}
      \label{alg:update}
    \end{algorithm}

      The strength of the conjugate-exponential representation is that both
      the prior and posterior take the form of an exponential-family
      distribution over $\boldeta$ that can be summarized by succinct
      hyperparameters $\nu$ and $\bchi$.
      \begin{align}\label{eqn:exponential-posterior}
	P(\boldeta|\bchi,\nu) &= \tilde{h}(\boldeta)\exp\left(
        \boldeta^{\intercal}\bchi - \nu A(\boldeta) -
	\tilde{A}(\bchi,\nu)\right)
      \end{align}

      \begin{figure*}[t]
	\centering
	\includegraphics{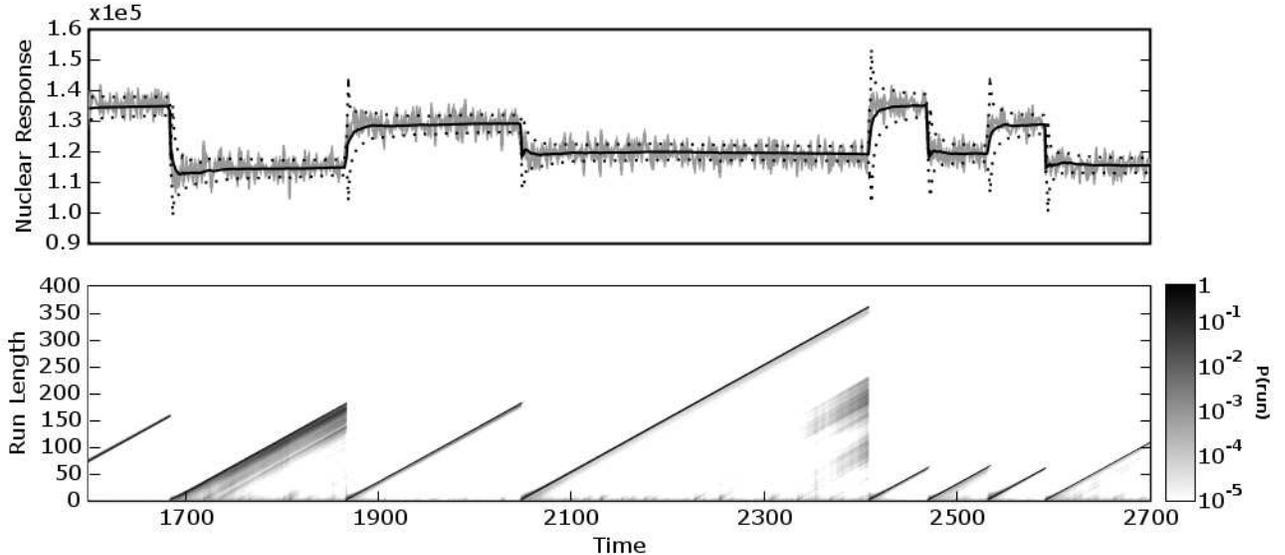}
	\vskip -0.5cm
	\caption{The top plot is a 1100-datum subset of nuclear magnetic
	response during the drilling of a well.  The data are plotted in
	light gray, with the predictive mean (solid dark line) and
	predictive 1-$\sigma$ error bars (dotted lines) overlaid.  The
	bottom plot shows the posterior probability of the current run
	$P(r_{t}\given \bx_{1:t})$ at each time step, using a logarithmic
	color scale.  Darker pixels indicate higher probability.}
	\label{fig:well-log}
      \end{figure*}
 
      We wish to infer the parameter vector $\boldeta$ associated with the
      data from a current run length $r_{t}$.  We denote this run-specific
      model parameter as $\boldeta_{t}^{(r)}$.  After finding the posterior
      distribution $P(\boldeta_{t}^{(r)} \given r_{t},\bx_{t}^{(r)})$, we
      can marginalize out the parameters to find the predictive
      distribution, conditional on the length of the current run.
      \begin{align}\label{eqn:exponential-predictive}
	P(x_{t+1} \given r_{t}) &=
	\int d\boldeta\; P(x_{t+1}\given\boldeta)
	               P(\boldeta_{t}^{(r)}\shorteq\boldeta \given
		        r_{t},\bx_{t}^{(r)})
      \end{align}

      This marginal predictive distribution, while generally not itself an
      exponential-family distribution, is usually a simple function of the
      sufficient statistics.  When exact distributions are not available,
      compact approximations such as that described by Snelson and
      Ghahramani \cite{snelson-ghahramani-2005a} may be useful.  We will
      only address the exact case in this paper, where the predictive
      distribution associated with a particular current run length is
      parameterized by $\nu_{t}^{(r)}$ and $\bchi_{t}^{(r)}$.
      \begin{align}
	\label{eqn:posterior-nu}
	\nu_{t}^{(r)} &= \nu_{\sf{prior}} + r_{t}\\
	\label{eqn:posterior-chi}
	\bchi_{t}^{(r)} &= \bchi_{\sf{prior}}
	  + \sum_{t'\in r_{t}}\boldsymbol{u}(x_{t'})
      \end{align}
    
    \subsection{COMPUTATIONAL COST}
      The complete algorithm, assuming exponential-family likelihoods, is
      shown in Algorithm \ref{alg:update}.  The space- and time-complexity
      per time-step are linear in the number of data points so far
      observed.  A trivial modification of the algorithm is to discard the
      run length probability estimates in the tail of the distribution
      which have a total mass less than some threshold, say $10^{-4}$.
      This yields a constant average complexity per iteration on the order
      of the expected run length $E[r]$, although the worst-case complexity
      is still linear in the data.

      \begin{figure*}[t]
	\centering
	\includegraphics{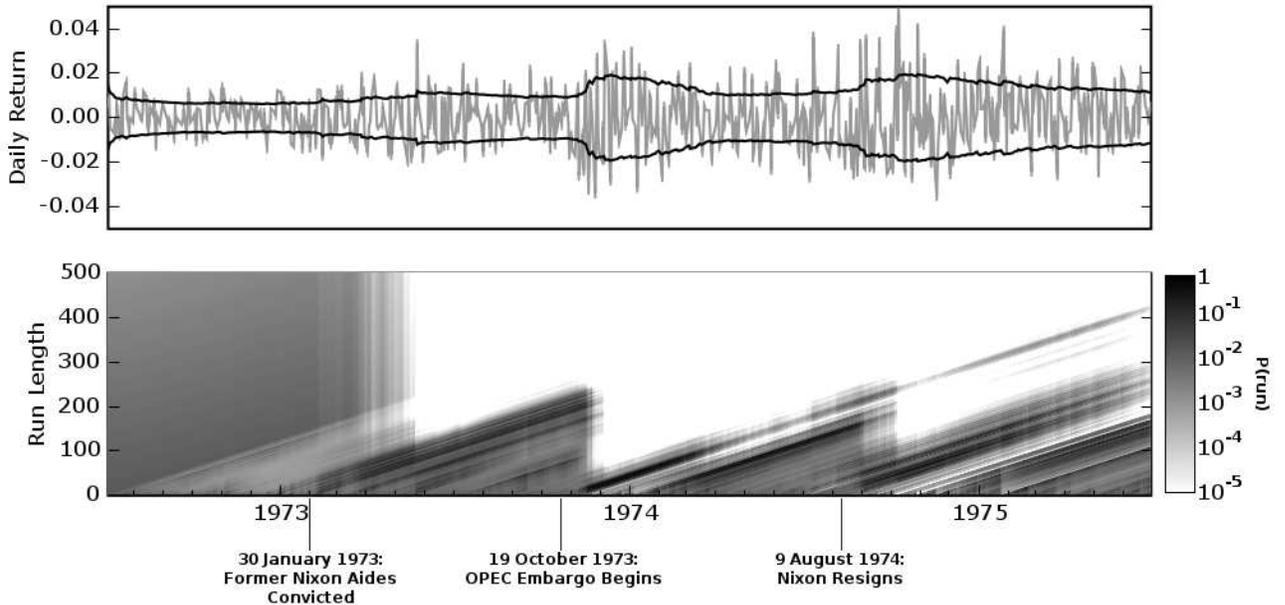}
	\vskip -0.5cm
	\caption{The top plot shows daily returns on the Dow Jones
	  Industrial Average, with an overlaid plot of the predictive
	  volatility.  The bottom plot shows the posterior probability of
	  the current run length $P(r_{t}\given \bx_{1:t})$ at each time
	  step, using a logarithmic color scale.  Darker pixels indicate
	  higher probability.  The time axis is in business days, as this
	  is market data.  Three events are marked: the conviction of
	  G. Gordon Liddy and James W. McCord, Jr. on January 30, 1973; the
	  beginning of the OPEC embargo against the United States on
	  October 19, 1973; and the resignation of President Nixon on
	  August 9, 1974.}
	\label{fig:watergate-djia}
      \end{figure*}
   
  \section{EXPERIMENTAL RESULTS}
    In this section we demonstrate several implementations of the
    changepoint algorithm developed in this paper.  We examine three
    real-world example datasets.  The first case is a varying Gaussian mean
    from well-log data.  In the second example we consider abrupt changes
    of variance in daily returns of the Dow Jones Industrial Average.  The
    final data are the intervals between coal mining disasters, which we
    model as a Poisson process.  In each of the three examples, we use a
    discrete exponential prior over the interval between changepoints.

    \subsection{WELL-LOG DATA}
      These data are 4050 measurements of nuclear magnetic response taken
      during the drilling of a well.  The data are used to interpret the
      geophysical structure of the rock surrounding the well.  The
      variations in mean reflect the stratification of the earth's crust.
      These data have been studied in the context of changepoint detection
      by \'{O} Ruanaidh and Fitzgerald \cite{oruanaidh-fitzgerald-1996a},
      and by Fearnhead and Clifford \cite{fearnhead-clifford-2003a}.

      The changepoint detection algorithm was run on these data using a
      univariate Gaussian model with prior parameters $\mu = \sn{1.15}{5}$
      and $\sigma = \sn{1}{4}$.  The rate of the discrete exponential
      prior, $\lambda_{\sf{gap}}$, was 250.  A subset of the data is
      shown in Figure \ref{fig:well-log}, with the predictive mean and
      standard deviation overlaid on the top plot.  The bottom plot shows
      the log probability over the current run length at each time step.
      Notice that the drops to zero run-length correspond well with the
      abrupt changes in the mean of the data.  Immediately after a
      changepoint, the predictive variance increases, as would be expected
      for a sudden reduction in data.
    
    \subsection{1972-75 DOW JONES RETURNS}
      During the three year period from the middle of 1972 to the middle of
      1975, several major events occurred that had potential macroeconomic
      effects.  Significant among these are the Watergate affair and the
      OPEC oil embargo.  We applied the changepoint detection algorithm
      described here to daily returns of the Dow Jones Industrial Average
      from July 3, 1972 to June 30, 1975.  We modelled the returns
      \begin{align}
	R_{t} &= \frac{p^{\sf{close}}_{t}}{p^{\sf{close}}_{t-1}}-1,
      \end{align}
      (where $p^{\sf{close}}$ is the daily closing price) with a zero-mean
      Gaussian distribution and piecewise-constant variance.  Hsu
      \cite{hsu-1977a} performed a similar analysis on a subset of these
      data, using frequentist techniques and weekly returns.

      We used a gamma prior on the inverse variance, with $a=1$ and
      $b=10^{-4}$.  The exponential prior on changepoint interval had rate
      $\lambda_{\sf{gap}}=250$.  In Figure \ref{fig:watergate-djia}, the
      top plot shows the daily returns with the predictive standard
      deviation overlaid.  The bottom plot shows the posterior probability
      of the current run length, $P(r_{t}\given \bx_{1:t})$.  Three events
      are marked on the plot: the conviction of Nixon re-election officials
      G. Gordon Liddy and James W. McCord, Jr., the beginning of the oil
      embargo against the United States by the Organization of Petroleum
      Exporting Countries (OPEC), and the resignation of President Nixon.

    \subsection{COAL MINE DISASTER DATA}
\begin{figure*}[t]
	\centering
	\includegraphics[width=\textwidth]{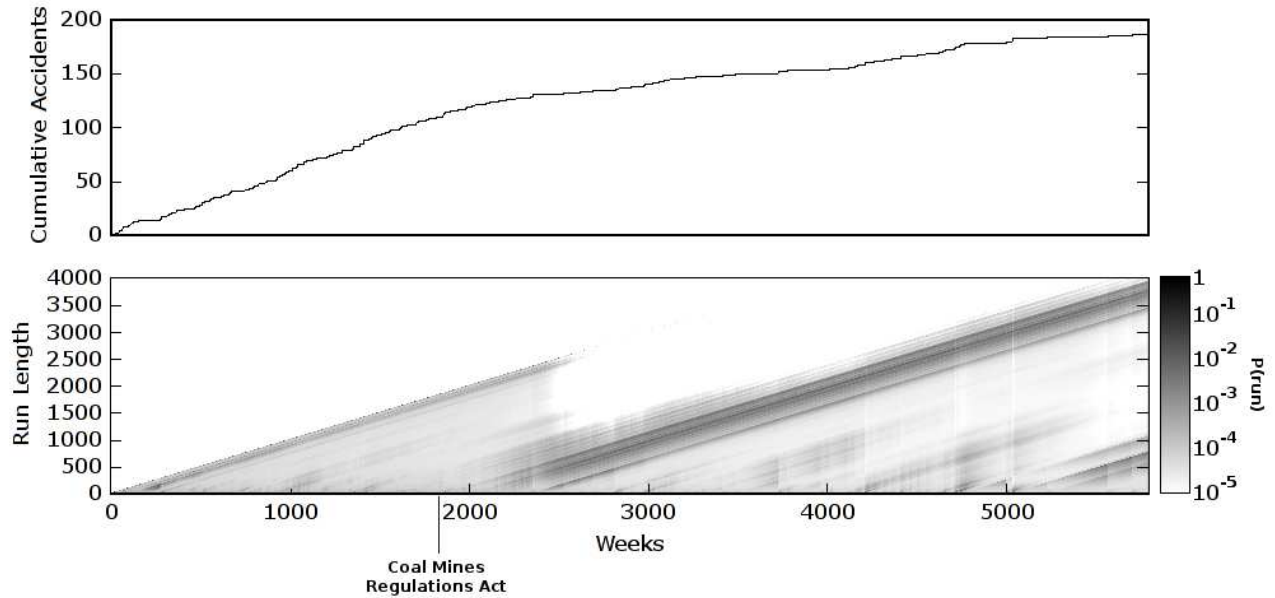}
	\vskip -0.4cm
	\caption{These data are the weekly occurrence of coal mine
	disasters that killed ten or more people between 1851 and 1962.
	The top plot is the cumulative number of accidents.  The accident
	rate determines the local average slope of the plot.  The
	introduction of the Coal Mines Regulations Act in 1887 is marked.
	The year 1887 corresponds to weeks 1868 to 1920 on this plot.  The
	bottom plot shows the posterior probability of the current run
	length at each time step, $P(r_{t}\given \bx_{1:t})$.}
	\label{fig:coal-mines}
      \end{figure*}

      These data from Jarrett \cite{jarrett-1979a} are dates of coal mining
      explosions that killed ten or more men between March 15, 1851 and
      March 22, 1962.  We modelled the data as an Poisson process by weeks,
      with a gamma prior on the rate with $a=b=1$.  The rate of the
      exponential prior on the changepoint inverval was
      $\lambda_{\sf{gap}}=1000$.  The data are shown in Figure
      \ref{fig:coal-mines}.  The top plot shows the cumulative number of
      accidents.  The rate of the Possion process determines the local
      average of the slope.  The posterior probability of the current run
      length is shown in the bottom plot.  The introduction of the Coal
      Mines Regulations Act in 1887 (corresponding to weeks 1868 to 1920)
      is also marked.

  \section{DISCUSSION}
    This paper contributes a predictive, online interpetation of Bayesian
    changepoint detection and provides a simple and exact method for
    calculating the posterior probability of the current run length.  We
    have demonstrated this algorithm on three real-world data sets with
    different modelling requirements.

    Additionally, this framework provides convenient delineation between
    the implementation of the changepoint algorithm and the implementation
    of the model.  This modularity allows changepoint-detection code to use
    an object-oriented, ``pluggable'' type architecture.
    
  \subsubsection*{Acknowledgements}
    The authors would like to thank Phil Cowans and Marian Frazier for
    valuable discussions.  This work was funded by the Gates Cambridge
    Trust.

  \bibliography{changepoint}
  \bibliographystyle{plain}
  
\end{document}